\documentclass[letterpaper, conference, dvipsnames]{ieeeconf}
\IEEEoverridecommandlockouts
\usepackage[utf8]{inputenc}
\usepackage{xargs} 
\usepackage{tabularx}
\usepackage{amssymb,amsmath}

\usepackage{amsthm}
\usepackage{epsfig}
\usepackage{textcomp} 
\usepackage{epstopdf}
\usepackage{url}
\usepackage{cite}
\usepackage{wrapfig}
\usepackage{marvosym}  
\usepackage{enumerate}
\usepackage{soul}
\usepackage{tipa}
\usepackage[percent]{overpic}
\usepackage{tikz}
\usetikzlibrary{arrows.meta}
\usepackage{pgfgantt}
\usepackage[linesnumbered,ruled,vlined]{algorithm2e}
\usepackage[textsize=footnotesize]{todonotes}
\usepackage{graphbox}
\usepackage{microtype} 

\newtheorem{problem}{Problem}

\theoremstyle{definition}

\theoremstyle{remark}

\newcommand{\changed}[1]{\textcolor{black}{#1}}

\usepackage{xspace}
\def\ddms{\textbf{\texttt{DDM}}\xspace}

\title{
DDM: Fast Near-Optimal Multi-Robot Path Planning using Diversified-Path\\
and Optimal Sub-Problem Solution Database Heuristics
}
\author{
Shuai D. Han \quad Jingjin Yu
\thanks{
S. D. Han and J. Yu are with the Department of Computer Science, 
Rutgers, the State University of New Jersey, Piscataway, NJ, USA. E-Mails: 
\{{\tt shuai.han, jingjin.yu}\}\hspace*{.25em}\MVAt \hspace*{.25em}cs.rutgers.edu. 
}
}

\begin{document}

\maketitle

\begin{abstract}

We propose a novel centralized and decoupled algorithm, \ddms, for 
solving multi-robot path planning problems in grid graphs,
targeting on-demand and automated warehouse-like settings. 
Two settings are studied: a traditional one whose objective is to move 
a set of robots from their respective initial vertices to the goal vertices 
as quickly as possible, and a dynamic one which requires frequent 
re-planning to accommodate for goal configuration adjustments. 
Among other techniques, \ddms is mainly enabled through exploiting two 
innovative heuristics: path diversification and optimal sub-problem 
solution databases. The two heuristics attack two distinct phases of 
a decoupling-based planner: while path diversification allows 
the more effective use of the entire workspace for robot travel, optimal 
sub-problem solution databases facilitate the fast resolution of local 
path conflicts. Extensive evaluation demonstrates that \ddms achieves 
high levels of scalability and high levels of solution optimality. 

\end{abstract}

\section{Introduction}\label{sec:introduction}
Labeled optimal multi-robot path planning (MPP) problems, despite their 
high associated computational complexity~\cite{YuLav13AAAI}, have been 
actively studied for decades due to the problems' extensive applications. The 
general task is to efficiently plan high-quality, collision-free 
paths to route a set of robots from an initial configuration to a goal 
configuration. 
Traditionally, the focus of studies on MPP is mainly with {\em one-shot} 
problems where the initial and goal configurations are pre-specified, 
and both are equal in cardinality to the number of the robots. More 
recently, an alternative {\em dynamic} formulation has started to attract 
more attention due to its real-world relevance \cite{WurDanMou08}. 
A dynamic instance keeps assigning new goals to robots that already 
reached their current goals, thus requiring algorithms that can actively 
re-plan the paths to accommodate adjustments of goal configuration. 

In this paper, we propose the \ddms (\textbf{D}iversified-path 
\textbf{D}atabase-driven \textbf{M}ulti-robot Path 
Planning) algorithm, capable of quickly computing near-optimal 
solutions to large-scale labeled MPPs, under both one-shot and dynamic 
settings on grid graphs. At a high-level, adapting the classic and 
effective decoupled planning paradigm \cite{ErdLoz86,sanchez2002using,
bennewitz2002finding,BerOve05,BerSnoLinMan09}, 
\ddms first generates a shortest path between each pair of start 
and goal vertices and then resolves local conflicts among the initial 
paths. In generating the initial paths, a {\em path diversification} 
heuristic is introduced that attempts to make the path ensemble use 
all graph vertices in a balanced manner, which minimizes the chance 
that many robots aggregate in certain local areas, causing unwanted 
congestion. Then, in resolving path conflicts, we observe that most 
conflicts can be resolved in a local $2 \times 3$ or $3 \times 3$ area. 
Based on the observation, a second novel heuristic is introduced which 
builds a {\em \changed{min-makespan} solution database for all $2 \times 3$ and 
$3\times 3$ sub-problems}, and ensures quick local conflict resolution 
via database retrievals. Together, the two heuristics produce 
simultaneous improvement on both computational efficiency and solution 
optimality in terms of computing near-optimal solutions under practical 
settings, as compared with state-of-the-art methods, e.g., \cite{YuLav16TRO}. 
For example, our algorithm 
can compute $1.x$ optimal solutions for a few hundreds of 
robots on a $60 \times 60$ grid with $10\%$ obstacles in under a second. 

{\bf Related Work.} 
MPP has been actively studied for many decades~\cite{Gol84, ErdLoz86, 
LavHut98b, GuoPar02}, which is perhaps mainly due to its hardness 
and simultaneously, its practical importance. Both one-shot and dynamic 
MPP formulations find applications in a wide range of domains 
including evacuation~\cite{RodAma10}, 
formation~\cite{PodSuk04, SmiEgeHow08}, localization~\cite{FoxBurKruThr00}, 
microdroplet manipulation~\cite{GriAke05}, object 
transportation~\cite{RusDonJen95}, search and rescue~\cite{JenWheEva97}, 
human robot interaction~\cite{knepper2012pedestrian}, and large-scale 
warehouse automation~\cite{WurDanMou07,WurDanMou08}, to list a few. 
\changed{MPP is known as Multi-Agent Pathfinding (MAPF)~\cite{stern2019multi}.}

In the past decade, significant progress has been made on 
solving one-shot MPP problems. Optimal \changed{and sub-optimal} solvers are achieved through 
reduction to other problems, e.g., SAT~\cite{Sur12}, answer set 
programming~\cite{erdem2013general}, and network flow~\cite{YuLav16TRO}. 
Decoupled approaches \cite{ErdLoz86}, which first compute independent 
paths and then try to avoid collision afterward, are also popular.
Commonly found decoupled approaches in a graph-based setting include 
independence detection~\cite{StaKor11}, sub-dimensional 
expansion~\cite{wagner2015subdimensional} and conflict-based 
search~\cite{boyarski2015icbs, cohen2016improved}. Similar to our 
approach, there is a decoupled algorithm~\cite{wang2011mapp} which 
uses online calculations over local graph structures to handle path 
interactions. However, \cite{wang2011mapp} only explored simple 
local interactions without much consideration to optimality. 
There also exists prioritized methods~\cite{BerOve05, bennewitz2002finding, 
saha2006multi,BerSnoLinMan09} and a divide-and-conquer 
approach~\cite{yu2018constant} which achieve decent scalability but 
at the cost of either completeness or optimality. 
\changed{
Some anytime algorithms~\cite{vedder2019x} are proposed to 
quickly find a feasible solution and then improve it. 
}
A learning-assisted approach~\cite{sigurdson2019automatic} has recently 
been developed to automatically pick the algorithm that is likely to perform 
well on a given MPP task. 

Dynamic MPP with new goals appearing over time, although not as 
extensively studied as its one-shot counterpart, has started to receive 
more attention. The problem is particularly applicable to automated 
warehouse systems~\cite{WurDanMou08}. Recent work has focused on the
dynamic warehouse MPP setup, pursuing both better planning 
algorithms~\cite{ma2017lifelong} and robust execution 
schedules~\cite{hoenig2019persistent}. Prioritized planning method 
with a flexible priority sequence has also been developed~\cite{okumura2019priority}.

MPP is widely studied from many other perspectives. As such, our literature 
coverage here is necessarily limited; readers are referred to~\cite{TurMicKum14,
YuLav12CDC,SolHal12,blm-rvo,snape2011hybrid, ChiHanYu2018WAFR,
adler2015efficient, SolYu15} for some additional algorithmic developments 
on MPP under unlabeled (i.e., robots are indistinguishable), partially 
labeled, and continuous settings. 

The topic of path diversification has been explored under both single 
and multi-robot settings. For single robot exploring a domain with many 
obstacles, obtaining a path ensemble can increase the chance of succeeding
in finding a longer horizon plan~\cite{branicky2008path,
erickson2009survivability}. Similar to what we observe in the current 
study, path diversity is just one of the relevant factors affecting 
search success~\cite{knepper2009path}. Survivability is also examined 
under a probabilistic framework for multi-robot systems \cite{lyu2016k}.
\changed{In a similar context, a heuristic based on path conflicts expediates 
the solution process of an MPP algorithm~\cite{barer2014suboptimal}.}

Finally, the use of a sub-problem solution database {\em trades off between  
offline and online computation}, which is a general principle that finds frequent 
applications in robotics, e.g., \cite{tedrake2009lqr,hauser2017learning}.
Relating to MPP, a similar technique called {\em pattern database} has 
been used in solving large $(n^2 - 1)$-puzzles~\cite{felner2005solving,
felner2007compressed}, as well as problems like Sokoban~\cite{pereira2013finding}. 

\textbf{Main Contributions.} 
This work brings three main contributions. 
First, based on the insight that decoupled MPP solvers tend to generate 
individual paths that aggregate in certain local areas (e.g., center of 
the workspace), we introduce path diversification heuristics that make 
more effective uses of the entire workspace. 
Second, the $2 \times 3$ and $3 \times 3$ sub-problem optimal solution 
databases, constructed one-time-only for resolving local path conflicts, 
bring significant on-line computational savings. 
Lastly, the first two main contributions jointly yield the \ddms 
algorithm, which is effective not only for one-shot settings but also for
dynamic MPP problems, as demonstrated through our extensive evaluation 
efforts. 

\textbf{Scope.} We explicitly point out that \ddms targets 
structured warehouse-like environments. As such, \ddms is not suitable for
MPPs with narrow passages, which remains challenging to be effectively 
solved. The current work focuses on synchronous path generation and does 
not address the equally important path execution aspects. Nevertheless,
\ddms can be readily combined with path execution approaches, e.g., 
\cite{hoenig2019persistent}, to form a complete planning and execution pipeline. 

\textbf{Organization.} 
In Section~\ref{sec:preliminaries}, we formally define both the one-shot and 
dynamic MPP formulations, and introduce assumptions. 
In Section~\ref{sec:algorithm}, we provide an overview of \ddms. 
In Section~\ref{sec:initial-path} and Section~\ref{sec:3x3}, we describe the 
path diversification heuristics and the sub-problem solution database, respectively. 
In Section~\ref{sec:experiments}, we provide \changed{evaluation} results of \ddms. 
We conclude in Section~\ref{sec:conclution}.

\section{Preliminaries}\label{sec:preliminaries}

\subsection{One-shot Multi-Robot Path Planning}\label{sec:mrp}
Consider $n$ robots in an undirected grid graph $G(V, E)$. 
Given integers $w$ and $h$ as the width and height of the grid, 
we define the vertex set of $G$ as 
$V \subseteq \{(i, j) | 1 \leq i \leq w, 1 \leq j \leq h\}$; the 
elements not in $V$ are considered as static obstacles. Following the 
traditional $4$-way connectivity rule, for each vertex $(i, j) \in V$, its 
neighborhood is $N(i) = \{(i + 1, j), (i - 1, j), (i, j + 1), (i, j - 1)\} 
\cap V$. For a robot $i$ with initial and goal vertices $x^I_i, x^G_i \in 
V$, a {\em path} is defined as a sequence of $T + 1$ vertices $P_i = (p^0_i, \dots, 
p^T_i)$ satisfying: 
{\em (i)} $p^0_i = x^I_i$; 
{\em (ii)} $p^T_i = x^G_i$; 
{\em (iii)} $\forall 1 \leq t \leq T$, $p^{t - 1}_i = p^t_i$ or $p^{t - 1}_i\in N(p^t_i)$. 
Denoting the joint initial and goal 
configurations of the robots as $X^I = \{x_1^I, \dots, x_n^I\} \subseteq V$ and 
$X^G = \{x_1^G, \dots, x_n^G\} \subseteq V$, the path set of all the robots 
is then $\mathcal P = \{P_1, \dots, P_n\}$.

For $\mathcal P$ to be collision-free, 
$\forall 1 \leq t \leq T$, $P_i, P_j \in \mathcal P$ must satisfy: 
{\em (i)} $p_i^t \neq p_j^t$ (no conflicts on vertices); 
{\em (ii)} $(p_i^{t - 1}, p_i^t) \neq (p_j^t, p_j^{t - 1})$ (no ``head-to-head''
collisions on edges). 

An optimal solution minimizes the {\em makespan} $T$, which is the time for all 
the robots to reach the goal vertices. 

\begin{problem}{\bf Time-optimal Multi-robot Path Planning (MPP).}\label{prob:mpp}
    Given $\langle G, X^I, X^G\rangle$, find a collision-free path set $\mathcal P$ 
    that routes the robots from $X^I$ to $X^G$ and minimizes $T$.
\end{problem}

\subsection{Dynamic Multi-Robot Path Planning}

The dynamic MPP formulation inherits most of one-shot MPP's structure, but with a 
few key differences. First, a robot $i$ will be assigned a new goal 
vertex when reaching its current goal $x^G_i$. Such a new goal 
is sampled from $V \backslash X^G$ using a certain distribution. Note that 
$X^G$ is continuously updated as new goals are assigned to the robots. Second, 
the optimization criteria is changed since the problem has no specific end state. 
In this paper, we maximize {\em system throughput}, which is the average number of 
goal arrivals in a unit of time. 

\begin{problem}{\bf Dynamic Multi-robot Path Planning (DMP).}\label{prob:dmp}
    Given $\langle G, X^I, X^G\rangle$, route the robots in $G$, accommodate for 
    changes in $X^G$, and maximize the system throughput.
\end{problem}

\subsection{Assumptions on the Graph Structure}

\begin{figure}[ht!]
    \centering
    \includegraphics[align = c, width = 0.99 \linewidth]
    {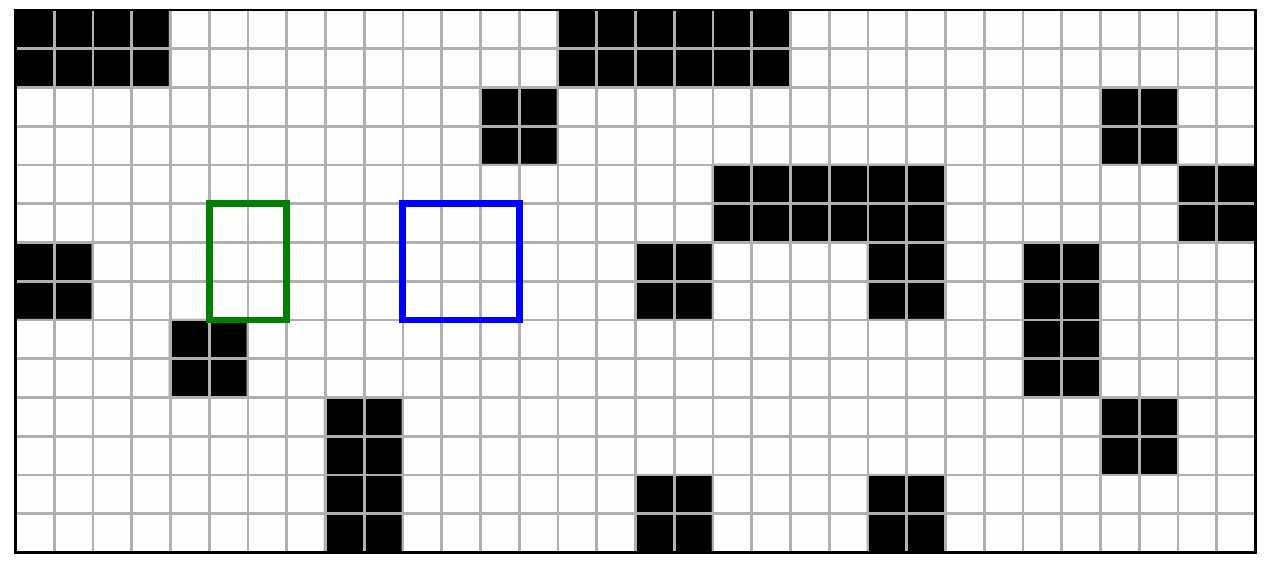}
    \caption{\label{fig:low-resolution-graph}
    \changed{A $k = 2$ low-resolution graph with $20\%$ obstacles. 
    The white cells are vertices, and the black cells are obstacles. 
    The green and blue rectangles visualize a $2 \times 3$ and a $3 \times 3$ sub-graph. 
    }}
\end{figure}

\changed{
In this paper, we assume that the graph $G$ does not contain narrow 
passages, and the width of a passage in $G$ is at least $k \geq 2$. 
Formally speaking, we define $G$ as a {\em low resolution graph}: 
given $k$ as the narrowest passage width, there exists a bijection between the set of all 
grid Graphs $\mathcal G$ to the set of all low resolution graphs 
$\mathcal G_{\text{low}}^k$: 
$\mathcal G \rightarrow \mathcal G_{\text{low}}^k, G(V, E) \mapsto 
G_{\text{low}}^k(V_{\text{low}}^k, E_{\text{low}}^k)$, where 
$V_{\text{low}}^k = \{(k i + x, k j + y) \mid \forall 
(i, j) \in V, x, y \in \{0, \dots, k - 1\}\}$.
A low-resolution graph example is provided in 
Fig.~\ref{fig:low-resolution-graph}. By the definition, $k = 2$ and $k = 3$ are sufficient 
to ensure that all vertices in $G$ are contained in some $2 \times 3$ or  
$3 \times 3$ sub-graphs, respectively. 
}
The restriction on 
low-resolution graphs effectively prevents environments with narrow passages
and mimics typical warehouse environments~\cite{WurDanMou08}.

\changed{
An essential component of our approach is the routing of robots inside 
some {\em obstacle-free} $2 \times 3$ and $3 \times 3$ sub-graphs. 
Examples of these sub-graphs are provided in Fig.~\ref{fig:low-resolution-graph}. 
With the problem setup introduced in Section~\ref{sec:mrp}, 
an MPP sub-problem in such a local sub-graph is always feasible~\cite{yu2018constant}, 
even when the sub-graph is fully occupied by robots.}

Since \ddms pipeline remains the same when using $2 \times 3$ or 
$3 \times 3$ sub-graphs, in the following sections we only use the $3 \times 3$ 
sub-graph structure to introduce \ddms.

\section{Overview of the \ddms Algorithm}\label{sec:algorithm}
\begin{figure}[ht!]
    \centering
    \begin{overpic}[keepaspectratio, width = 0.99 \linewidth]
        {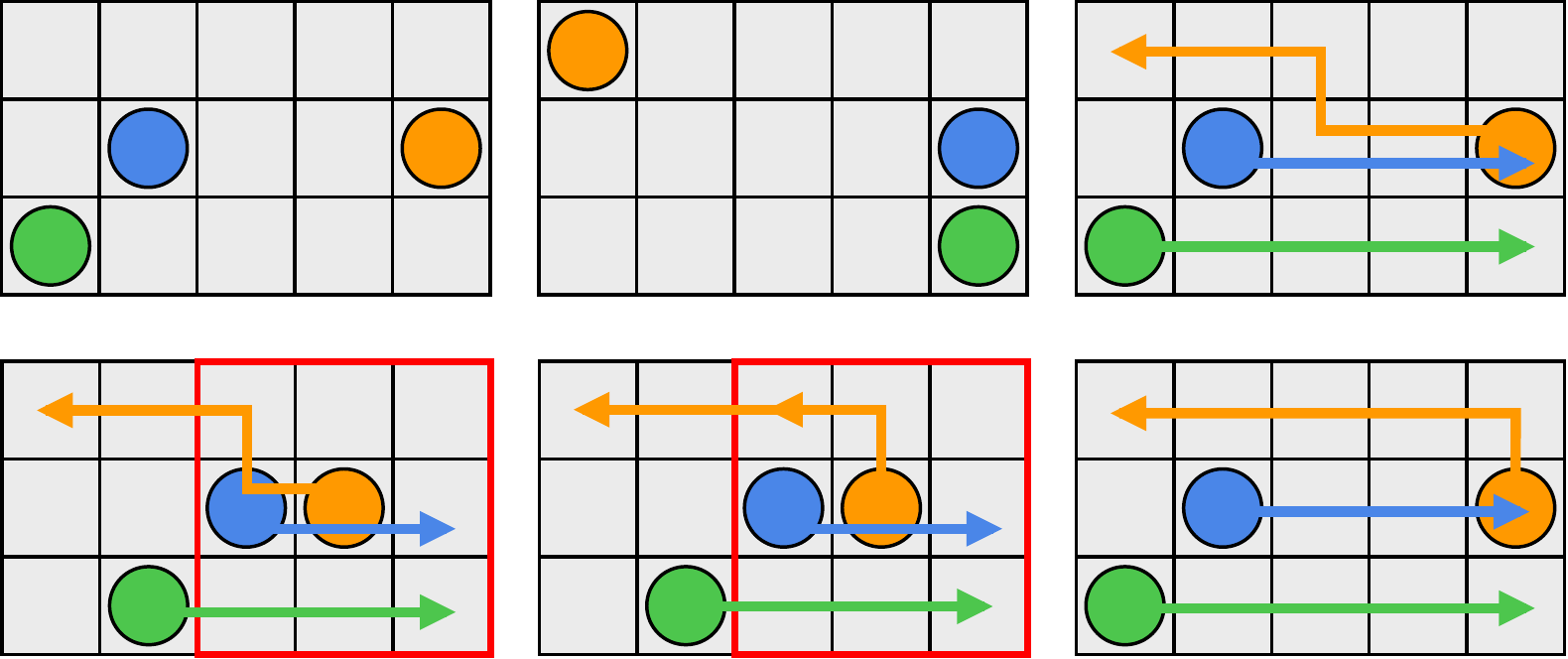}
        \footnotesize
        \put(13.55, 20.5){(a)}
        \put(48,    20.5){(b)}
        \put(82,    20.5){(c)}
        \put(13.55, -2.5){(d)}
        \put(48,    -2.5){(e)}
        \put(82,    -2.5){(f)}
    \end{overpic}
    \caption{\label{fig:algorithm-pipeline}
    Illustration of the \ddms solution pipeline. 
    (a) In a $5 \times 3$ graph, the initial configuration of $3$ robots are 
    visualized using blue, orange, and green disks. 
    (b) The goal configuration. 
    (c) The individual paths between each initial and goal vertex pairs are 
    visualized using arrowed lines. 
    (d) After \changed{simulating} the individual paths for one time step, we find a collision 
    in the next time step between the blue and orange robots. At this time, a local 
    $3 \times 3$ sub-graph is assigned to these two robots to resolve the conflict. 
    The boundary of the sub-graph is highlighted in red. 
    (e) The paths are updated using the $3 \times 3$ solution. The robots are able 
    to execute their new paths and get to the goal configuration without 
    colliding with each other. 
    (f) Alternative initial paths that are collision-free, which can be generated 
    by using the path diversification heuristics. 
    }
\end{figure}
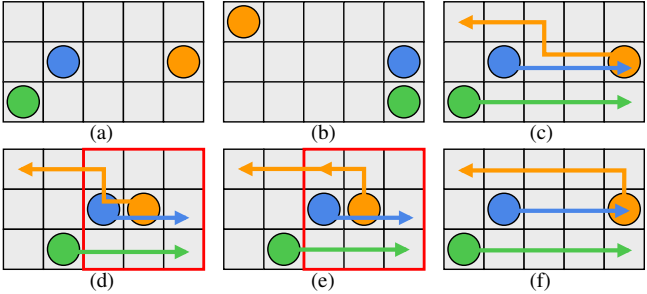

\ddms follows the decoupled paradigm and first creates a shortest path 
for each robot from its initial vertex to goal vertex, ignoring other robots. 
Then, a simulated execution is carried out. As conflicts are 
detected, they are resolved within {\em local} sub-graphs. 
\changed{An illustration of the \ddms pipeline is provided in 
Fig.~\ref{fig:algorithm-pipeline}. }
Although 
\ddms is described as a centralized method, the conflict resolution 
phase can be readily decentralized. This is especially applicable to 
DMP: after the initial paths are acquired, collision avoidance can be 
implemented locally; during the path execution stage, any robot may change 
its desired path without causing a system failure since conflict resolution 
is performed on the fly. 

Algorithm~\ref{alg:main} describes \ddms. 
In line~\ref{alg:main:init_var}, two structures are initialized: 
\changed{
$X^C$ which keeps track of robots'current locations, 
and $\mathcal G_{3 \times 3}$ which keeps a record of currently occupied 
$3 \times 3$ sub-graphs used for collision avoidance. 
}

Then, in line~\ref{alg:main:initial_path}, \ddms plans a shortest path from 
$x^I_i$ to $x^G_i$ for each robot $i$, without considering any interactions 
with the other robots. The detailed initial path generation process and its 
optimization techniques (i.e. the path diversification heuristics) are discussed 
in Section~\ref{sec:initial-path}. 

After the initial paths are acquired, \ddms starts 
\changed{to carry out a simulated execution of these paths and 
resolves the conflicts between them. }
At the beginning of each simulation time step, 
\ddms first checks whether collisions will occur if the robots all move along 
their planned paths (line~\ref{alg:main:next_step}--\ref{alg:main:conflicting_pair}). 
If no collision is detected, the collision avoidance procedures are skipped and the 
pipeline enters the execution stage 
(line~\ref{alg:main:execution_3x3}--\ref{alg:main:remove_3x3}). 

\begin{algorithm}[t]
    \small
    \DontPrintSemicolon
    $X^C \gets X^I$, $\mathcal G_{3 \times 3} \gets \varnothing$\; 
    \label{alg:main:init_var}
    $\mathcal P_{\text{planned}} = \{P_1, \dots, P_n\} \gets $ {\sc GetPaths}($G, X^I, X^G$)\; 
    \label{alg:main:initial_path}
    \While{$X^C \neq X^G$}{ \label{alg:main:execution_begin}
        $X^N \gets $ {\sc GetNextStep}($\mathcal P_{\text{planned}}$)\; 
        \label{alg:main:next_step}
        \For{$(i, j) \in$ {\sc CollidingRobotPairs}($X^C, X^N$)}{ 
            \label{alg:main:conflicting_pair}
            \If{$G_{3 \times 3} \gets$ {\sc FindSubGraph}($G, X^C, i, j, \mathcal G_{3 \times 3}$)}{
                \label{alg:main:find3x3}
                $R \gets \{i \, | \, x_i^C \in G_{3 \times 3}, \forall x_i^C \in X^C\}$\;
                \label{alg:main:find_robots}
                $X^I_{3 \times 3} \gets \{x^C_i \, | \, i \in R\}$ \;  
                $X^G_{3 \times 3} \gets $
                {\sc TempGoals}($G_{3 \times 3}, R, \mathcal P_{\text{planned}}$)\; 
                \label{alg:main:assign_sg}
                $\mathcal P_{\text{planned}} \gets $ 
                {\sc CheckDatabase}($X^I_{3 \times 3}, X^G_{3 \times 3}$) \; 
                \label{alg:main:solve_3x3}
                $\mathcal G_{3 \times 3} \gets 
                \mathcal G_{3 \times 3} \cup \{G_{3 \times 3}$\} \; 
                \label{alg:main:update_3x3}
            }
        }
        $X^C, \mathcal P_{\text{planned}} \gets$ {\sc Simulate3x3Paths}($X^C, \mathcal P_{\text{planned}}, \mathcal G_{3 \times 3}$)\; 
        \label{alg:main:execution_3x3}
        $X^C, \mathcal P_{\text{planned}} \gets$ {\sc SimulateOtherPaths}($X^C, \mathcal P_{\text{planned}}$)\; 
        \label{alg:main:execution_others}
        $\mathcal G_{3 \times 3} \gets$ 
        {\sc RemoveOutdatedSubGraphs}($\mathcal G_{3 \times 3}$)\; 
        \label{alg:main:remove_3x3}
    }
    \caption{\changed{Centralized \ddms for one-shot MPP}}\label{alg:main}
\end{algorithm}

When collisions occur, \ddms enters the collision avoidance stage 
(line~\ref{alg:main:conflicting_pair}--\ref{alg:main:update_3x3}). 
\changed{
Here, whenever we {\em iterate through} robots, 
we give robots further away from their goal configurations 
a higher priority for the potential decrease of global makespan.}

\begin{figure}[ht!]
    \centering
    \begin{overpic}[keepaspectratio, width = 0.99 \linewidth]{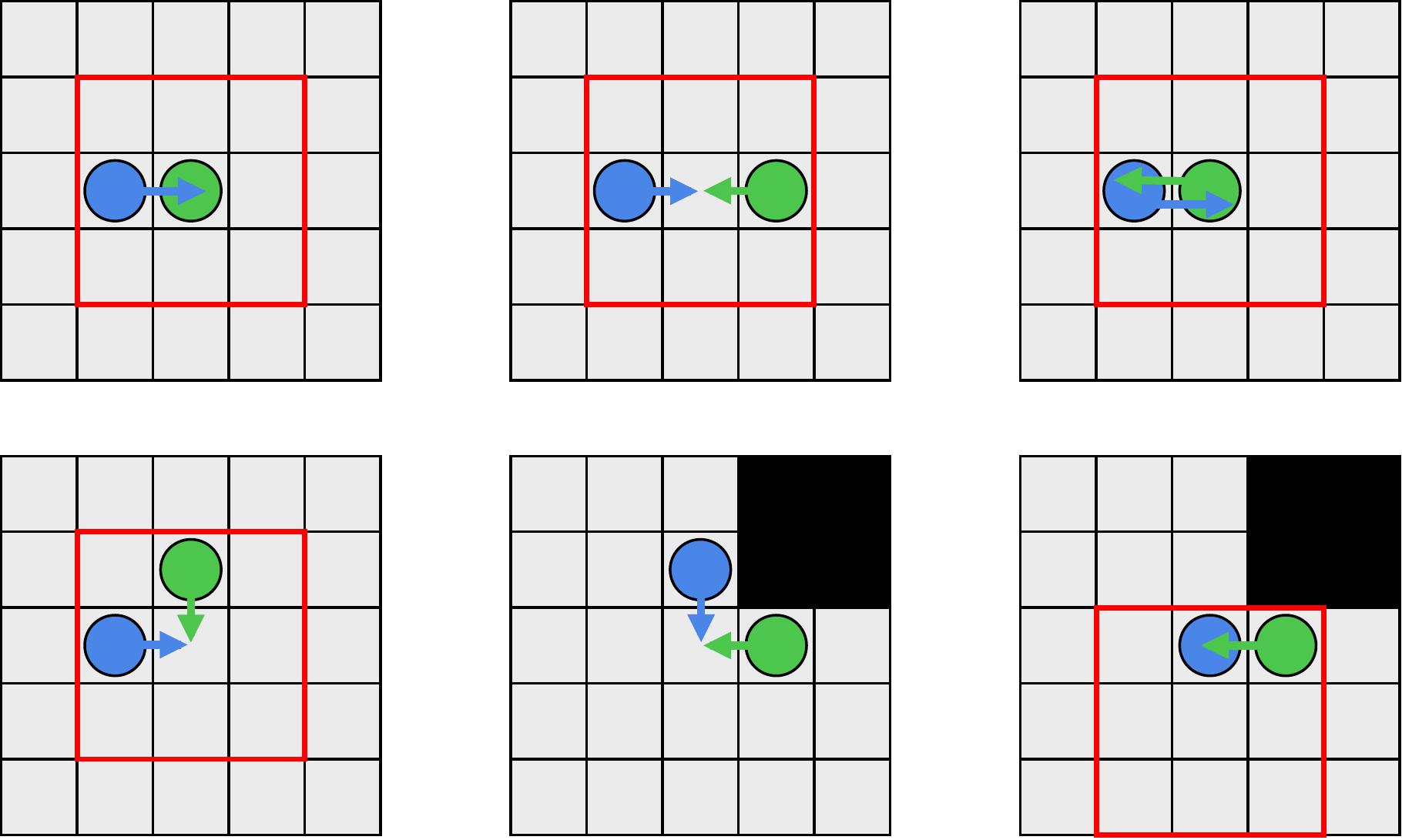}
        \footnotesize
        \put(11.7, 29.2){(a)}
        \put(48, 29.2){(b)}
        \put(84.5,   29.2){(c)}
        \put(11.7, -3.5){(d)}
        \put(48, -3.5){(e)}
        \put(84.5,   -3.5){(f)}
    \end{overpic}
    \caption{\label{fig:collision-types}
    \changed{
        (a--d) All types of collisions, including collisions on a vertex and 
        head-to-head collisions on an edge, can be contained in some 
        $3 \times 3$ graphs (drawn in red). 
        (e) The only scenario that we cannot find a $3 \times 3$ sub-graph due to 
        the black obstacle. 
        (f) We can find a $3 \times 3$ graph for the scenario in 
        sub-figure (e) by postponing one robot's move. }}
\end{figure}
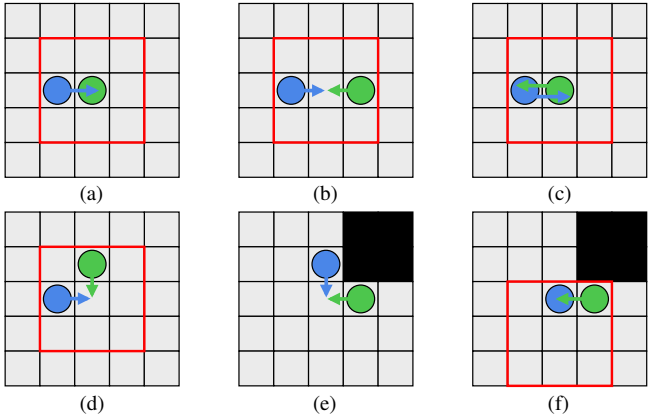

In line~\ref{alg:main:find3x3}, 
\changed{we {\em iterate through} all the pairs of conflicting robots 
and for each pair of them,} \ddms first 
attempts to find an \changed{\em obstacle-free} $3 \times 3$ sub-graph which meets two requirements: 
{\em (i)} it contains both conflicting robots. 
{\em (ii)} it does not overlap with currently occupied $3 \times 3$ graphs 
in $\mathcal G_{3 \times 3}$. 
\changed{
For requirement {\em (i)}, when $G$ is obstacle-free, we can always find a 
$3 \times 3$ graph that covers the colliding robots. 
Examples of such $3 \times 3$ sub-graphs for all collision types 
are provided in Fig.~\ref{fig:collision-types}(a-d). 
Note that since the $3 \times 3$ sub-graphs shown in the sub-figures are 
not the only choices, when $G$ is a $k = 3$ low-resolution graph, we can still 
find $3 \times 3$ sub-graphs that meet requirement {\em (i)}, except for the only 
outlier case shown in Fig.~\ref{fig:collision-types}(e). 
In this case, we {\em postpone} the conflict by letting one robot wait 
and the other robot move. 
After one step of simulation, a $3 \times 3$ graph satisfies requirement {\em (i)} 
will be available (see Fig.~\ref{fig:collision-types}(f)). 
Note that a $3 \times 3$ sub-graph satisfies requirement {\em (i)} may not 
meet requirement {\em (ii)}. If we cannot find a $3 \times 3$ graph that 
satisfies both requirements, we skip this pair of conflicting robots and 
start to process the next pair. 
}

If a $3 \times 3$ sub-graph $G_{3 \times 3}$ is acquired, in 
line~\ref{alg:main:find_robots}, \ddms locates all the robots that are currently 
inside $G_{3 \times 3}$, whose paths will be affected by the conflict resolution process. 
\ddms then assigns temporary goal configurations to all these robots (line 
\ref{alg:main:assign_sg}) and route them inside $G_{3 \times 3}$ by looking 
for the solution in the database (line~\ref{alg:main:solve_3x3}). 
\changed{
For the temporary goal assignment, we {\em iterate through} the affected 
robots and for each robot, we inspect its desired path backwards 
(i.e., from the goal to the robot's current location) and check 
if a vertex in the path also appears in $G_{3 \times 3}$. 
We try to assign the first vertex appearing in $G_{3 \times 3}$ as the robot's temporary goal. 
The purpose of such an assignment is to move robots closer to 
goals during collision avoidance. 
If the desired vertex is already assigned to another robot, we then opt for 
a random vertex in $G_{3 \times 3}$ that is not assigned to other robots, 
since the temporary goals of different robots cannot be identical. 
}
\changed{In line~\ref{alg:main:solve_3x3}, the min-makespan paths for routing these robots 
to the temporary goals} are readily found in the 
database; further details is provided in Section~\ref{sec:3x3}. 
The initial planned paths are updated 
according to the $G_{3 \times 3}$ solution. Note that we might call {\sc GetPaths} 
(in line~\ref{alg:main:initial_path}) for a robot in case the original path and the 
$3 \times 3$ solution cannot be simply concatenated due to a non-desirable 
temporary goal assignment. The final step of the collision avoidance stage is to 
put $G_{3 \times 3}$ into $\mathcal G_{3 \times 3}$ 
(line~\ref{alg:main:update_3x3}). 

\begin{figure}[ht!]
    \centering
    \begin{overpic}[keepaspectratio, width = 0.99 \linewidth]
        {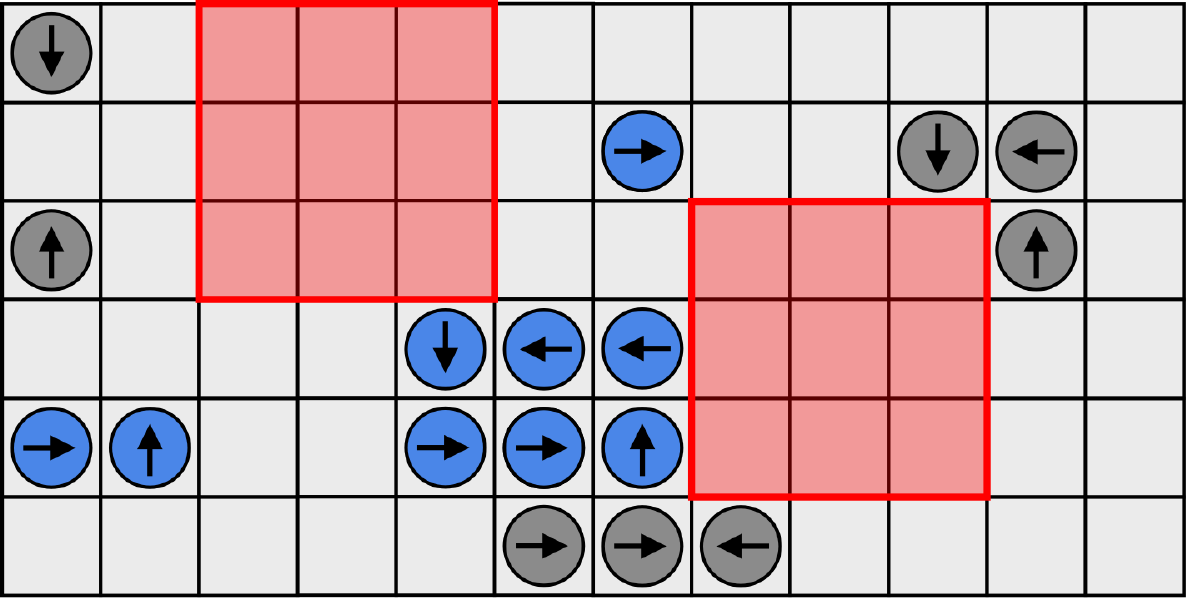}
        \footnotesize
    \end{overpic}
    \caption{\label{fig:execution-flow}
    Executing the planned paths which do not belong to $3 \times 3$ 
    sub-graph solutions. The red regions are the $3 \times 3$ regions 
    currently in $\mathcal G_{3 \times 3}$; robots in these regions are omitted. 
    The remaining robots are visualized using blue and gray disks, with arrows 
    indicating the desired next time step moves. A blue disk implies that the 
    robot is permitted to move, while a gray disk implies that the robot will 
    stay still. The robots at top-right are stopped since one of them 
    is trying to move into a sub-graph in $\mathcal G_{3 \times 3}$. The robots at 
    top-left and bottom are stopped due to collisions. We note 
    that the figure is only for illustrating purposes and do not reflect actual
    cases. }
\end{figure}

Recall that when constructing a $3 \times 3$ graph in line~\ref{alg:main:find3x3}, 
we require it not to overlap with other $3 \times 3$ graphs already in use, i.e., 
the elements in $\mathcal G_{3 \times 3}$. The requirement leaves some conflicts 
untreated, which are avoided in the following path execution process.  
In line~\ref{alg:main:execution_3x3}, the robots in the current occupied 
$3 \times 3$ graphs move first, since their paths are generated from the optimal 
solution database and are guaranteed to be 
collision-free. Then, in line~\ref{alg:main:execution_others}, we move the other 
robots while avoiding collisions between them: first, we find all the robots that 
are moving into the sub-graphs in $\mathcal G_{3 \times 3}$ and stop them, 
to avoid interruptions to the $3 \times 3$ solutions' execution; 
next, we detect collisions in the current step, and recursively stop all the robots 
that are involved in these collisions. 
An illustration of this path execution process is provided in 
Fig.~\ref{fig:execution-flow}. Finally, in line~\ref{alg:main:remove_3x3}, we 
remove elements from $\mathcal G_{3 \times 3}$ if we finished executing the 
corresponding $3 \times 3$ solutions. The untreated 
collisions mentioned in the beginning of this paragraph are either handled 
by the line~\ref{alg:main:execution_others}, or by constructing a $3 \times 3$ 
graph after the previous overlapping sub-graphs are removed from $\mathcal G_{3 \times 3}$. 

The performance of \ddms directly relates to the efficiency of the collision 
avoidance process, which is in turn influenced by {\em the total number of 
path conflicts} and {\em the time to resolve a conflict}. In the next two 
sections, we introduce optimization techniques including path 
diversification heuristics and the sub-graph solution database. These techniques 
enable \ddms to achieve high levels of scalability and solution quality.

\section{Path Diversification Heuristics}\label{sec:initial-path}

When individual paths are generated without care, their footprint tends 
to aggregate on portions of the graph environment, leading to higher 
chances for path conflicts. To alleviate this issue, multiple heuristics are 
attempted in this work. In the case where the graph is obstacle-free, we 
can reduce collisions by letting the robots go around the center. 
For graphs with arbitrary obstacles, we can reduce 
collisions by modifying the heuristic we used in the single robot path planning algorithm. 
As the number of potential collisions drops, \ddms can generate 
solutions that are closer to optimal. 

\subsection{Graph without Obstacles}

In an obstacle-free graph, the shortest path between two vertices is a set of 
axis-aligned moves according to the vertices' coordinate differences. 
For example, it takes $3$ steps along the x-axis and $2$ steps along the y-axis 
for a robot to move from 2D coordinate $(2, 3)$ to $(5, 5)$. Obtaining such a 
shortest path requires an ordering of these axis-aligned moves. Two ordering rules 
we studied are discussed as below. 

\begin{figure}[ht!]
    \centering
    \begin{overpic}[scale = 0.27, angle = 90]
        {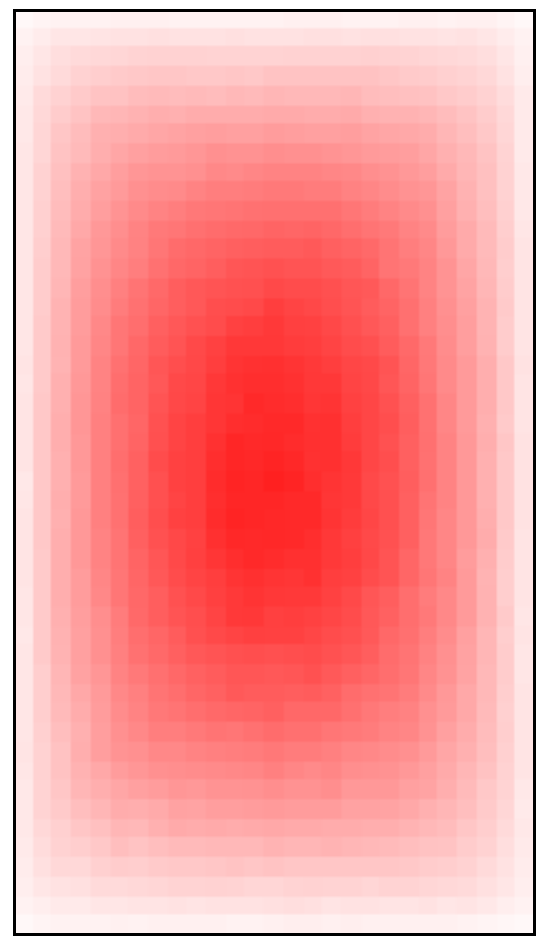}
        \footnotesize
        \put(45, -8.5){(a)}
    \end{overpic}\hspace{2mm}
    \begin{overpic}[scale = 0.27, angle = 90]
        {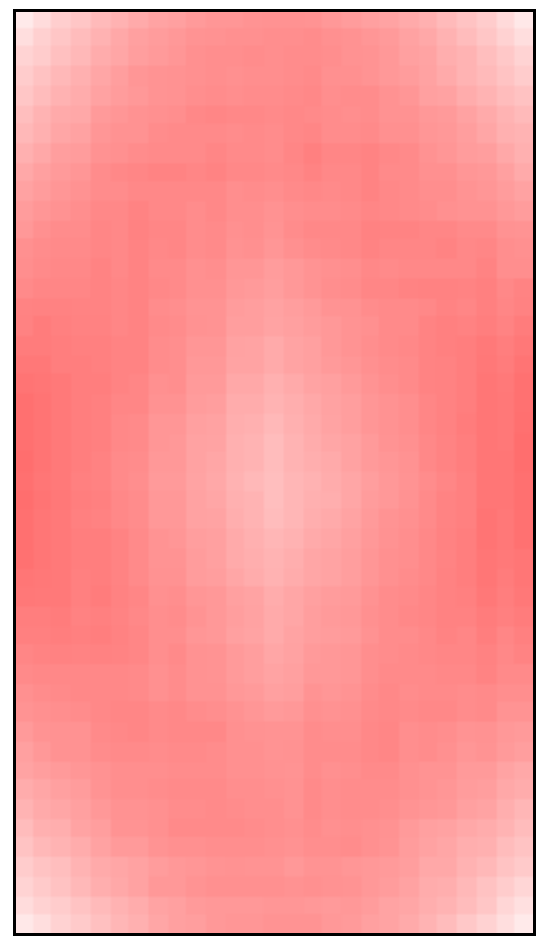}
        \footnotesize
        \put(45, -8.5){(b)}
    \end{overpic}\hspace{2mm}
    \begin{overpic}[scale = 0.27, angle = 90]
        {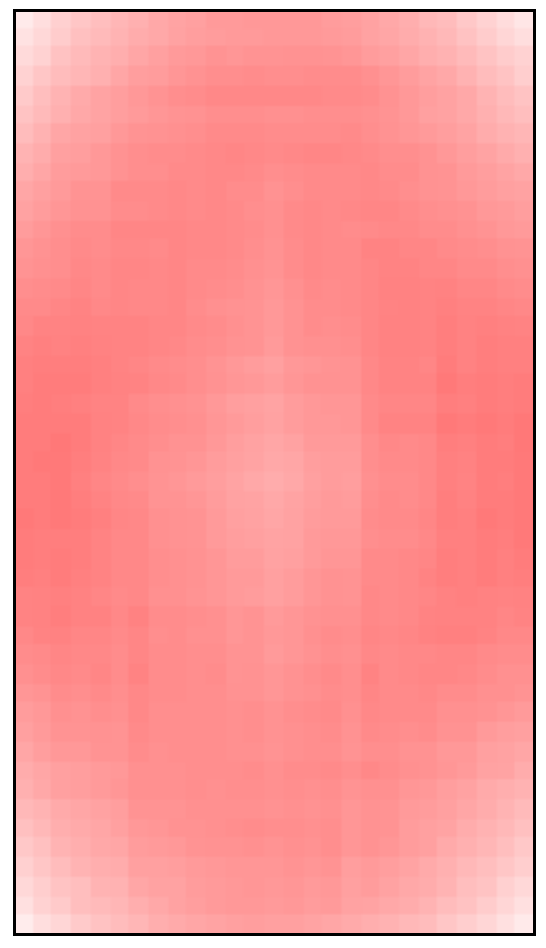}
        \footnotesize
        \put(45, -8.5){(c)}
    \end{overpic}
    \caption{\label{fig:initial-path-generation-no-obstacle}
    Comparison of different path-finding heuristics in an obstacle-free 
    $48 \times 27$ grid graph. We uniformly randomly sample $100000$ initial and 
    goal vertex pairs and generate the initial paths using the studied rules. The 
    color intensity of a cell reflects the number of time the cell is traversed by 
    a path: the darker the color, the heavier the cell is utilized. 
    (a) When using random paths, the center of the graph is congested. 
    (b) When using single-turn paths, we can avoid congestion in the center. 
    (c) A balance ($85\%, 15\%$) between two types of single-turn paths avoids 
    graph center under-utilization. 
    }
\end{figure}

{\bf Randomized Paths.} 
This baseline ordering rule returns a randomized sequence of the axis-aligned 
moves. That is, for a path consisting of $i$ moves along the x-axis and $j$ moves 
along the y-axis, the moving sequence is uniformly randomly picked from 
$\binom{i + j}{i}$ possible orderings. As evidenced by 
Fig.~\ref{fig:initial-path-generation-no-obstacle}(a), such a randomized sequence 
causes the graph center congested, and increases the initial paths' 
conflicts. 

{\bf Path Diversification Using Single-Turn Paths.} 
This heuristic moves a robot along one axis until the robot is aligned with the 
goal vertex, and then moves the robot along the other axis until it reaches the goal. 
There are two options when picking a single-turn path: depending on 
which axis to move along first, we can make the turning point closer or further 
away from the graph center. As indicated in 
Fig.~\ref{fig:initial-path-generation-no-obstacle}(b), we can avoid congestion in 
the center of the graph by always choosing the turning point that is further away 
from the center. 
In Fig.~\ref{fig:initial-path-generation-no-obstacle}(c), 
by mixing the selection of turning points that are closer and further 
away from the graph center, we can balance the vertex usage 
in the center and around the border. 

\subsection{Graph with Obstacles}
In a graph with obstacles, it is a natural choice to use the A* algorithm 
to generate the initial paths. The state space of the A* algorithm corresponds 
to the vertex set $V$. 

\begin{figure}[ht!]
    \centering
    \begin{overpic}[scale = 0.425, angle = 90]
        {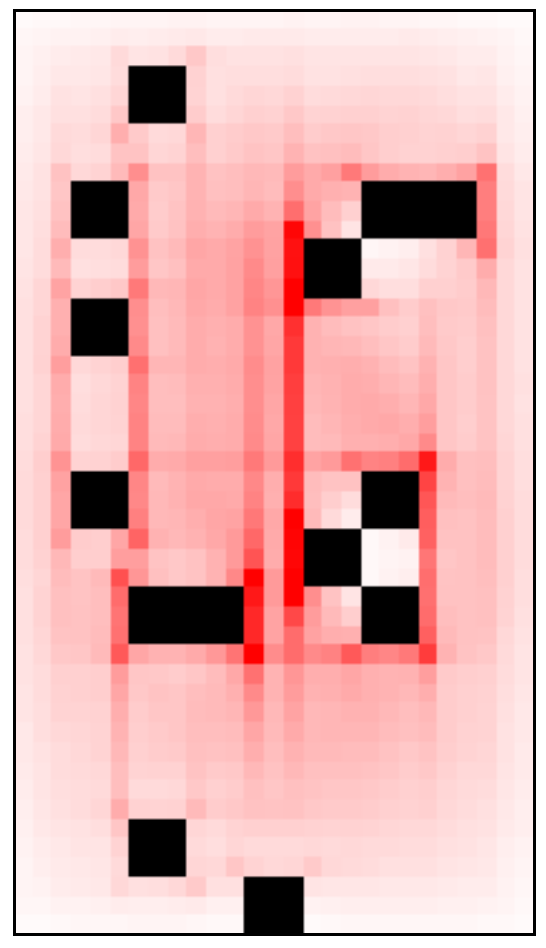}
        \footnotesize
        \put(45, -6){(a)}
    \end{overpic}
    \begin{overpic}[scale = 0.425, angle = 90]
        {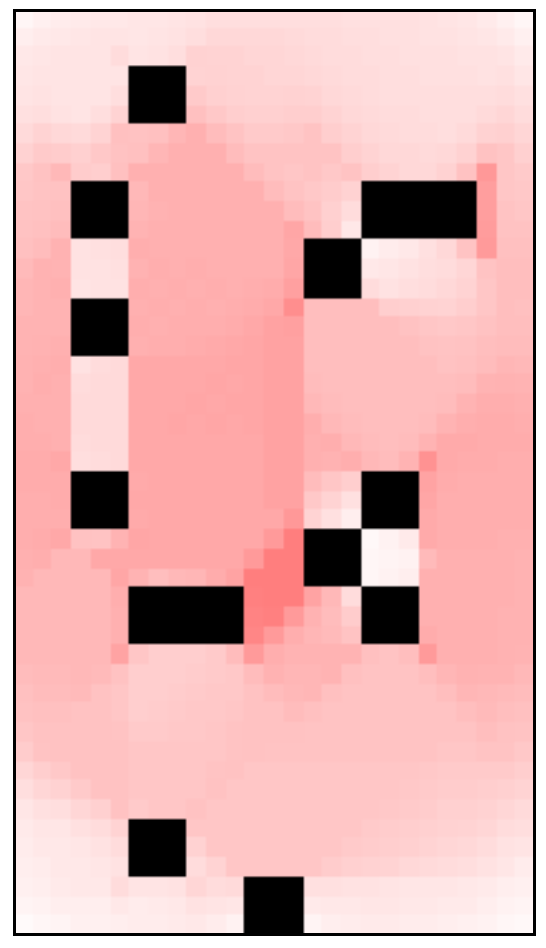}
        \footnotesize
        \put(45, -6){(b)}
    \end{overpic}
    \caption{\label{fig:initial-path-generation-with-obstacle}
    Comparison of different path finding heuristics in a graph with obstacles 
    visualized in black. Other visual elements are the same as the ones in 
    Fig.~\ref{fig:initial-path-generation-no-obstacle}. 
    (a) Pure Manhattan distance heuristic. 
    (b) Path diversification by vertex occupancy.  
    }
\end{figure}

{\bf The Manhattan Distance Heuristic.} 
As a well-known traditional heuristic for path planning on a grid, 
the Manhattan distance sums up the absolute differences of two points' Cartesian coordinates.
Given a goal vertex $(k, \ell)$, the heuristic value of a search state 
at vertex $(i, j)$ is calculated as 
\[H_{\text{Manhattan}}[(i, j)] = |i - k| + |j - \ell|.\]
The Manhattan distance heuristic leads to extensive utilization of vertices 
around obstacles and congestion on some high-traffic lanes 
(see Fig.~\ref{fig:initial-path-generation-with-obstacle}(a)). 

Since the initial path planning is performed sequentially over the $n$ robots, 
a path generated later can avoid conflicts with earlier paths. In this work, 
we realize this concept using two innovative path diversification heuristics. 

{\bf Path Diversification by Vertex Occupancy.} 
We define the {\em occupancy} of a vertex as the number of paths traverse through the 
vertex. Denoting $\mathbb N_0$ as the set of all non-negative 
integers, we construct a map $O: V \to \mathbb N_0$ to actively track the occupancy of 
all vertices throughout the initial path generation process. 
At the beginning, for each $(i, j) \in V$, $O[(i, j)] = 0$. 
Then, we sort the robots and generate initial paths for robots with goals further away 
(in terms of the Manhattan distance) from the initial vertices first. 
After each path is generated by the A* algorithm, $O$ is updated such that for each 
vertex $(i, j)$ in the path, $O[(i, j)]$ increments by $1$. 
The heuristic value for search state $(i, j)$ is calculated as 
\[H_{\text{Occupancy}}[(i, j)] = H_{\text{Manhattan}}[(i, j)] + O[(i, j)] / n.\]
Here, the last term of the equation refers to the additional cost 
imposed by path intersections. A constant value $n$ 
\changed{(i.e. the number of robots)}
is used to balance between 
finding a shorter path and finding a path with less interference with the others. 
In practice, we notice that this constant value ensures path diversification 
while keeps the initial paths short. 

Fig.~\ref{fig:initial-path-generation-with-obstacle}(b) demonstrates the effect 
of the vertex occupancy heuristic on a graph with obstacles. 
Comparing the two sub-figures, we observe reduced congestion around obstacles and 
high-traffic lanes when using the path diversification heuristic. 

{\bf Path Diversification by a State-Time Map.} 
As an alternative path diversification heuristic, instead of just calculating 
vertex usage, we also take the time domain into account. 
Since there are generally two types of collisions (see Fig.~\ref{fig:collision-types}(a)) 
between robots: on a vertex or head-to-head on an edge, we now store the state-time 
information in a map $S: (V \cup E, \mathbb N_0) \to \mathbb N_0$, which specifies the 
number of times a vertex or an edge is used at a certain time step. 
Now, for an A* search state at vertex $(i, j)$ with cost-to-go value $t$, its 
state-time heuristic value is calculated as 
\begin{align*}
    H_{\text{StateTime}}[(i, j), t] = & \, H_{\text{Manhattan}}[(i, j)] \\
    + ( S[(i, j), t] & + S[(\textsc{Parent}(i, j), (i, j)), t]) / n, 
\end{align*}
with the first term on the second line refers to path conflicts on vertices, and the 
second term refers to conflicts on edges. 

When comparing the vertex occupancy heuristic with the state-time heuristic, 
it is not hard to see that since the state-time heuristic takes the time domain 
into consideration, it generates initial path sets with less conflicts. 
However, due to the fact that the initial paths might be modified during the 
\ddms \changed{simulated execution} phase, as we will demonstrate in Section~\ref{sec:experiments}, 
the vertex occupancy heuristic provides overall better solutions in terms of makespan. 
This is as expected since the effect of the vertex occupancy 
heuristic is less likely to be affected by unsynchronized path execution.

\section{$2 \times 3$ and $3 \times 3$ Problem Solution Database}\label{sec:3x3}
We now provide the details of the optimal sub-problem solution database, 
especially how the database is generated. 

Generating the $2 \times 3$ database and the $3 \times 3$ database follow 
largely similar steps. Here, we use the case of $3\times 3$ database to 
illustrate the necessary computation, which requires a bit more technical 
trickeries than the $2\times 3$ case. 
Let $X_n$ be the set of all configurations of $n$ 
($1 \leq n \leq 9$) robots in a $3 \times 3$ graph, bijections exist
between the set of all problem instances, the set of all solutions, and 
all pairs of initial and goal combinations $X_n \times X_n$. The solution
space has a size 
\[\sum_{n = 1}^{9}|X_n \times X_n| = \sum_{n = 1}^{9}|X_n|^2 = 
\sum_{n = 1}^{9}(\binom{9}{n} n!)^2 \approx 3 \times 10^{11}, \]
which is too large to compute and store. In this section, we introduce how this 
issue is resolved by exploiting symmetry.

{\bf Permutation Elimination.} 
Instead of exploring all possible combinations of $X^I$ and $X^G$, for each problem 
instance recorded in the database, we always sort $X^I$. Note that we can still 
find the solution for an arbitrary pair of $X^I$ and $X^G$: we 
first apply a permutation $\pi$ to both $X^I$ and $X^G$, such that $\pi(X^I)$ is 
sorted. Denoting $\mathcal P$ as the solution for $\pi(X^I)$ and $\pi(X^G)$ in the 
database, the solution for $X^I$ and $X^G$ is then $\pi^{-1}(\mathcal P)$. 
\changed{More details are provided at the end of this section.}
The size of the $3 \times 3$ database is now reduced to 
\[\sum_{n = 1}^{9}\binom{9}{n}|X_n| = 
\sum_{n = 1}^{9}(\binom{9}{n}^2 n!) \approx 1.7 \times 10^7.\]


{\bf Group Actions.} 
When generating the database, after we calculated a solution $\mathcal P$ for 
certain $X^I$, $X^G$, this solution can possibly be translated to the solutions  
for other related instances by taking the same {\em action} to $X^I$, 
$X^G$ and $\mathcal P$. 

In this work, we explore two types of actions. The first one, based on 
{\em rotational symmetry}, is to {\em rotate} all the 
configurations by the same degree. After this rotation, the processed $\mathcal P$ 
becomes the solution for the rotated $X^I$, $X^G$. 
Denoting $r$ as rotating a configuration clockwise by $90$ degrees, the set of 
all possible rotations is then $\{1, r, r^2, r^3\}$. Here, $1 = r^4$, which is 
interpreted as rotating a configuration by $360$ degrees, has no effect. 

The second action, based on {\em mirror symmetry}, is to {\em flip} a 
configuration by the vertical middle line of a $3 \times 3$ graph. The set 
of flip actions is denoted as $\{1, f\}$. 

Combining the two types of actions, we have 
\[\{1, r, r^2, r^3\} \times \{1, f\} = \{1, r, r^2, r^3, f, fr, fr^2, fr^3\}.\] 
Here, $fr^2$ is interpreted as flipping the configuration, and then rotate 
the flipped configuration clockwise by $180$ degrees. 

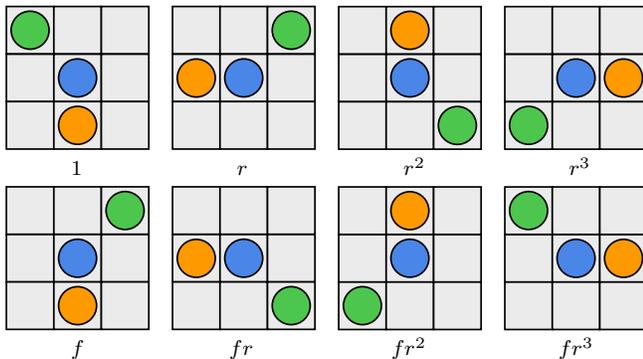
\begin{figure}[ht!]
    \centering
    \begin{overpic}[keepaspectratio, width = 0.99 \linewidth]{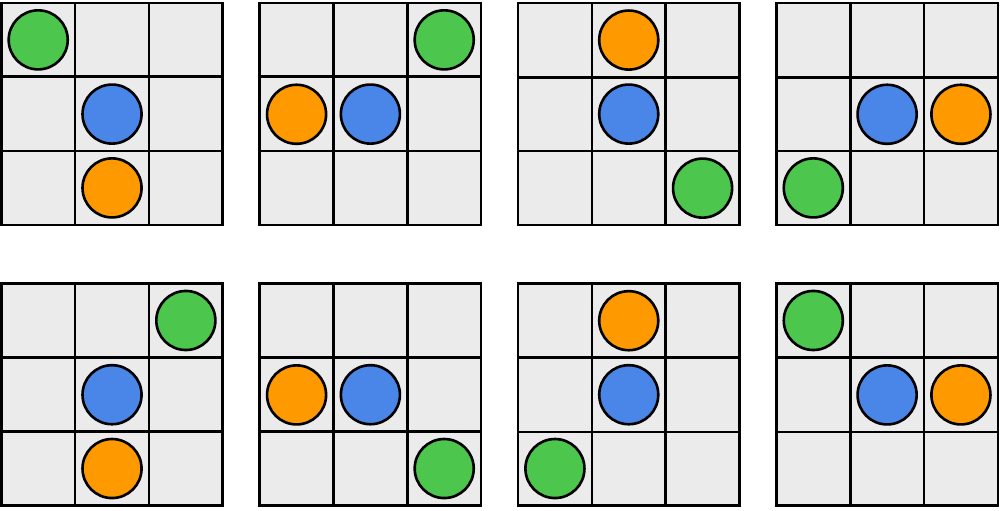}
        \footnotesize
        \put(10.2, 24.5){$1$}
        \put(36,   24.5){$r$}
        \put(61.9, 24.5){$r^2$}
        \put(87.6, 24.5){$r^3$}
        \put(10.2, -3.5){$f$}
        \put(34.8, -3.5){$fr$}
        \put(60,   -3.5){$fr^2$}
        \put(86,   -3.5){$fr^3$}
    \end{overpic}
    \caption{\label{fig:group-actions}
    Using group actions, we can generate up to eight different configurations out 
    of one configuration. 
    }
\end{figure}

With a set of calculated $X^I$, $X^G$, $\mathcal P$, 
applying each action in this set results in a new problem and the solution to it. 
Note that the group actions above already includes counter-clockwise rotations and 
other types of flipping. In Fig.~\ref{fig:group-actions}, we show the result of 
applying these actions to a configuration.

Moreover, $\mathcal P$ can be reversed to route the robots from $X^G$ to $X^I$. All 
in all, we can generate up to $16$ unique solutions out of the solution for a 
pair of $X^I$ and $X^G$ using this reversing process combined with group actions, 
\changed{which expedites the database generation process since we now only need to calculate around $1.1 \times 10^6$ problem instances. }

By permutation elimination and group actions, the computation time for obtaining 
the database is significantly reduced. Using an integer linear programming-based 
solver~\cite{YuLav16TRO}, generating the full $3 \times 3$ solution database takes 
about six hours. 

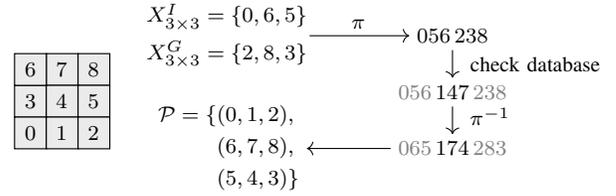
\begin{figure}[ht!]
    \centering
    \raisebox{0.5\height}{
    \begin{overpic}[keepaspectratio, width = 0.15 \linewidth]{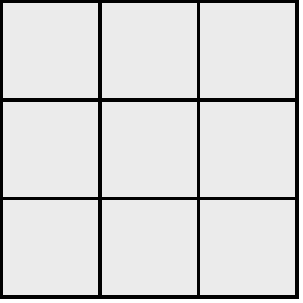}
        \footnotesize
        \put(11, 11){$0$}
        \put(43, 11){$1$}
        \put(76, 11){$2$}
        \put(11, 43){$3$}
        \put(43, 43){$4$}
        \put(76, 43){$5$}
        \put(11, 76){$6$}
        \put(43, 76){$7$}
        \put(76, 76){$8$}
    \end{overpic}}
    \hspace*{0.05in}
    \begin{tikzpicture}
        \footnotesize
        \node at (0, 1) {$X^I_{3 \times 3} = \{0, 6, 5\}$};
        \node at (0, 0.5) {$X^G_{3 \times 3} = \{2, 8, 3\}$};
        \node (key) at (3, 0.75) {$056\,238$};
        \node (val) at (3, 0) {\textcolor{gray}{$056$}$\,147\,$\textcolor{gray}{$238$}};
        \node (pval) at (3, -0.75) {\textcolor{gray}{$065$}$\,174\,$\textcolor{gray}{$283$}};
        \node (path) at (0, -0.75) {
            $\begin{aligned}
                \mathcal P = \{ & (0, 1, 2), \\
                & (6, 7, 8), \\
                & (5, 4, 3)\} \\
            \end{aligned}$};
        \node at (4.1, 0.375) {check database};
        \node at (3.5, -0.325) {$\pi^{-1}$};
        \node at (1.75, 0.9) {$\pi$};
        \draw[->] (1.1, 0.75) -- (key);
        \draw[->] (key) -- (val);
        \draw[->] (val) -- (pval);
        \draw[->] (pval) -- (path);
    \end{tikzpicture}
    \caption{\label{fig:database-lookup}\changed{
    An illustration of database lookup. 
    As demonstrated in the left figure, we give each vertex in the $3 \times 3$ 
    graph a unique id. 
    The example on the right is a database lookup for three robots. 
    First, a permutation $\pi$ is applied to sort the initial configuration from 
    $065$ to $056$; the goal configuration is updated accordingly. 
    After the transition steps between the sorted initial and goal configurations 
    are acquired from the database, a permutation $\pi^{-1}$ is applied to 
    generate a solution for the original problem. 
    }}
\end{figure}

\changed{
{\bf Database Lookup.} 
When we run \ddms, the database is pre-loaded into a C\texttt{++} STL map, 
with key and value both string types. 
Here, the key is a composition of the initial and goal configurations, 
and the value indicates the min-makespan paths between the two configurations. 
For a database lookup, we first apply a permutation to both the initial and 
goal configurations so that the initial configuration is sorted. 
Then, we compile the configurations into a single string and lookup for 
the value for this key in the database. 
Finally, we translate the value string into the solution paths. 
We provide an example of database lookup in Fig.~\ref{fig:database-lookup}. 
Our database is light-weight and fast to query: the $3 \times 3$ database uses 500MB 
disk storage, and takes 2GB memory when loaded into C\texttt{++} STL map; 
the $2 \times 3$ database is less than 300KB. Accessing $1000$ random 
keys sequentially takes {\em less than one millisecond} in total. 
}

\section{\changed{Simulation} Result}\label{sec:experiments}
\changed{
In this section, we compare \ddms with 
integer linear programming (ILP)~\cite{YuLav16TRO} and 
Enhanced Conflict-Based Search (ECBS)~\cite{barer2014suboptimal} 
under the classic one-shot MPP setting. 
The methods compared are to the best of our knowledge some of the 
fastest (near-)optimal solvers for MPP. 
For ILP, we evaluated the original optimal version {\em ILP Exact} and 
a sub-optimal variant {\em ILP $k$-way Split}. 
For ECBS, we set its weight parameter $w = 1.5$ since it seems to be   
a good balance between optimality and scalability in the 
original publication and from our observation. 
Besides MPP, we also tested \ddms on the dynamic formulation DMP. 
All our experiments are performed on an 
Intel\textsuperscript{\textregistered} Core\textsuperscript{TM} i7-6900k CPU at $3.2$GHz. 
Each data point is an average over $30$ runs on randomly generated instances. 
Although we do not provide a theoretical completeness and optimality 
guarantee for \ddms, the algorithm quickly solves all the problem 
instances we have tested, and provides near-optimal solutions.
}

\begin{figure}[ht!]
    \centering
    \includegraphics{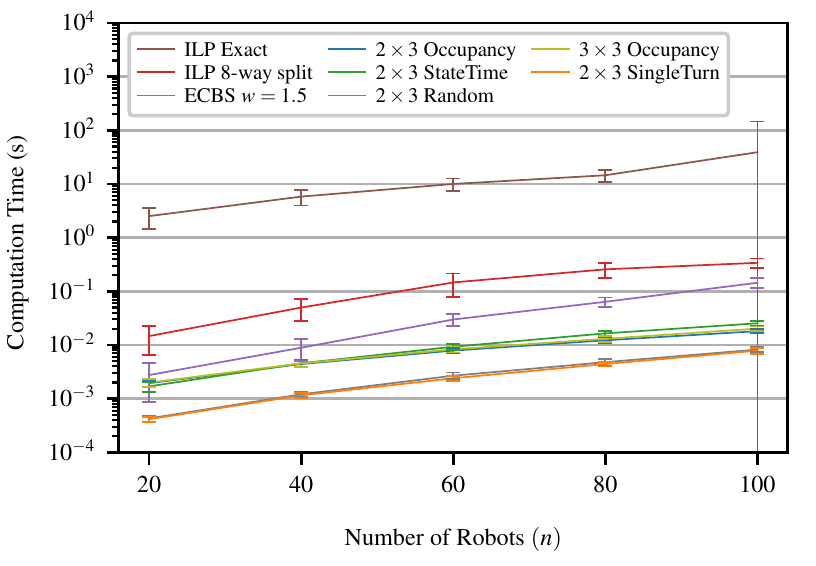}
    \includegraphics{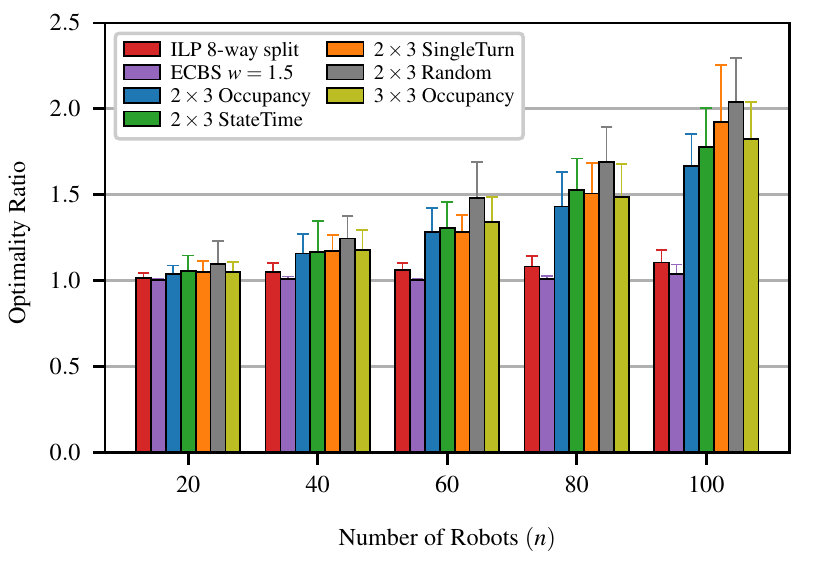}
	\caption{\label{fig:results-static-2418}
    Average computation time (top sub-figure) and makespan (bottom sub-figure), 
    \changed{and the standard deviations of algorithms on MPP tested in a 
    $24 \times 18$ obstacle-free grid, with varied number of robots. } 
    We use notation {\em Random} to indicate no path diversification heuristic is used. 
    } 
\end{figure}

We first examine a one-shot case on a $24 \times 18$ grid without obstacles. 
The start and goal vertices are uniformly randomly sampled. 
The result is compiled in Fig.~\ref{fig:results-static-2418}. 
Here, the $2 \times 3$ entires only construct $2 \times 3$ sub-graphs, 
and $3 \times 3$ entry means a $3 \times 3$ sub-problem is constructed 
whenever it is possible. 
The optimality ratio is calculated as the resulting makespan over the optimal 
makespan computed by ILP Exact. The comparison of computation time (the 
top sub-figure) shows that \ddms is the fastest method, which is about 
one to two magnitudes faster than the compared approaches.
In particular, $2\times 3$ SingleTurn is about $10^4$ times 
faster than ILP Exact and \changed{$60$} times faster than ECBS. 
At the same time, most of the \ddms variants maintained $1.x$ optimality. 

The result suggests that using $2 \times 3$ sub-graphs generates better 
solutions than using $3 \times 3$ ones, which is due to $2 \times 3$ 
graphs having a smaller footprint. Thus, interruptions to other robots 
is less likely. We hypothesized that resolving local conflicts using
$3 \times 3$ sub-graphs could help improve optimality; this turns 
out not to be the case in our tests. Nevertheless, for 
completeness, we include results on $3 \times 3$ sub-graphs. 

\begin{figure}[ht!]
    \centering
    \includegraphics[width = 0.99 \linewidth]{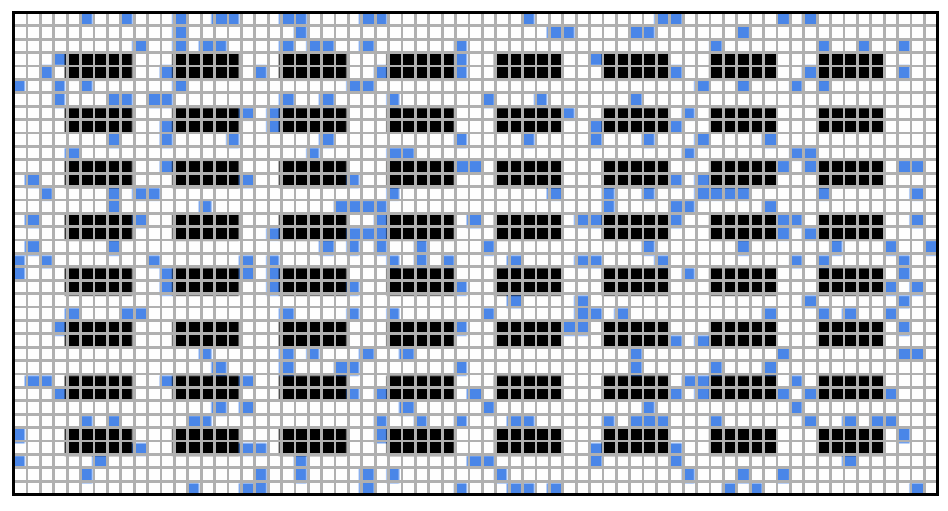}
    \caption{\label{fig:experiment-environments}
    A $69 \times 36$ warehouse-style workspace with $8$ row, $8$ column 
    $5 \times 2$ obstacle blocks. A random configuration of $300$ robots is colored in blue.
    }
\end{figure}
\begin{figure}[ht!]
    \centering
    \includegraphics{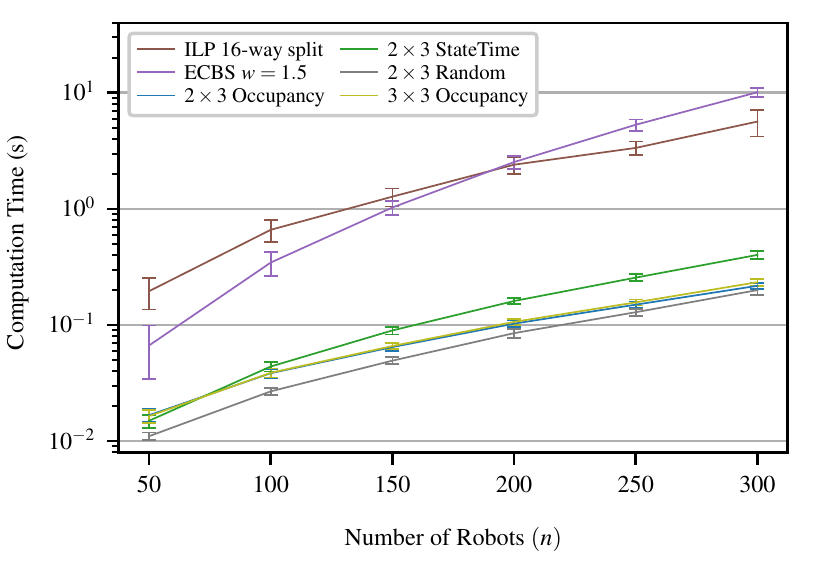}
    \includegraphics{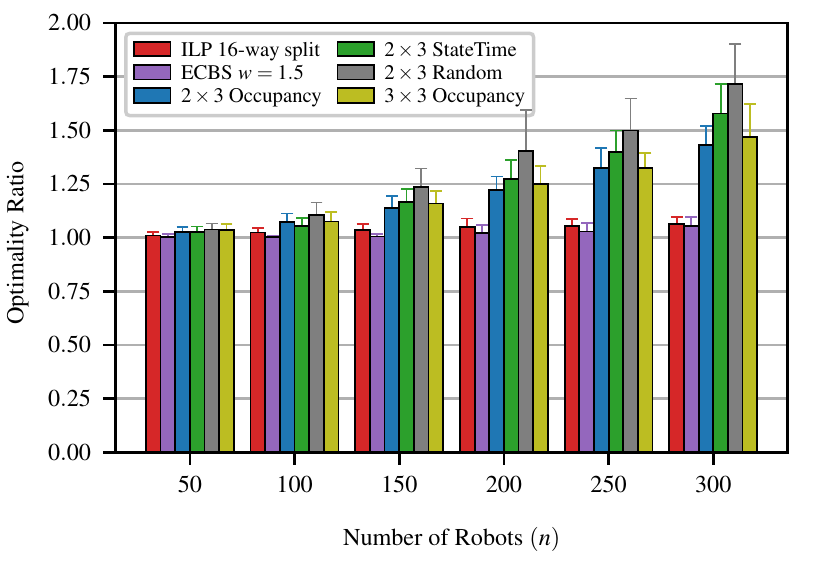}
    \caption{\label{fig:results-static-warehouse}
    Evaluation results of MPP in a warehouse-style workspace. 
    }
\end{figure}

In a second evaluation, we switch to a $69 \times 36$ warehouse-like 
environment (Fig.~\ref{fig:experiment-environments}) with many blocks of 
static obstacles. 
For this case, between $50$ and $300$ robots are 
attempted. The evaluation results are shown in 
Fig.~\ref{fig:results-static-warehouse}, which show similar performance 
trends as Fig.~\ref{fig:results-static-2418}. 
Here, because ILP Exact can no longer finish \changed{each and every calculation in ten minutes},  
comparison on optimality is made with respect to an underestimated 
makespan which is calculated without considering robot-robot collisions.
\ddms provides more than $50$ times speed up without sacrificing much 
optimality. 

\begin{figure}[ht!]
    \centering
    \includegraphics{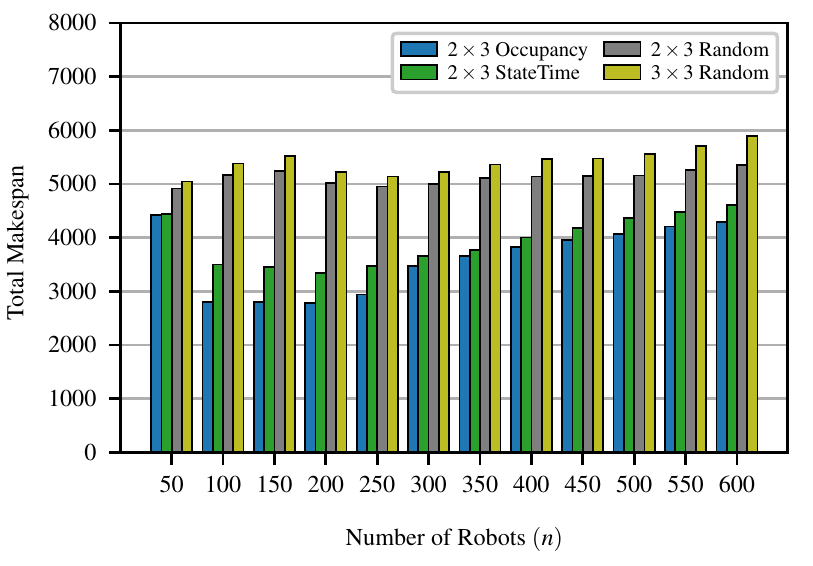}
    \includegraphics{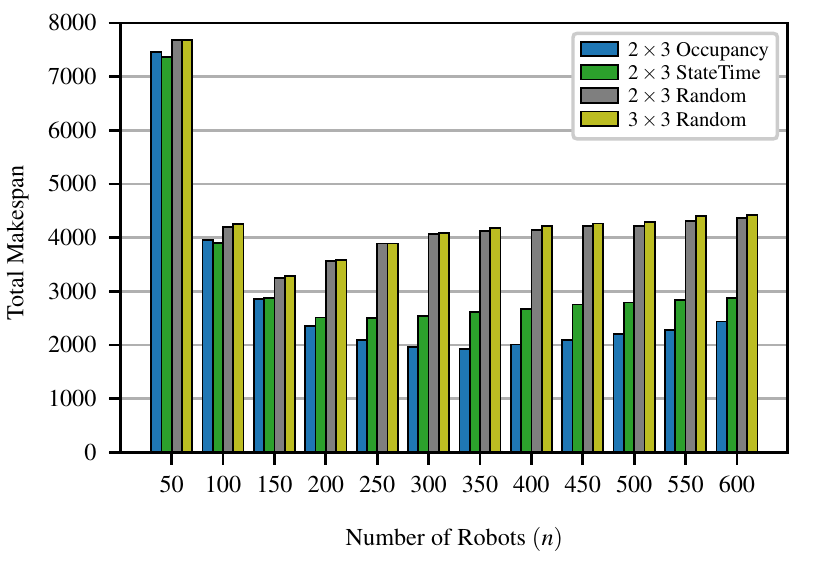}
    \caption{\label{fig:results-dynamic}
    Evaluation results of DMP in (top) a $30 \times 30$ low-resolution grid 
    ($k = 2$) with $10\%$ ($90$) obstacles, and (bottom) the warehouse-style 
    workspace in Fig.~\ref{fig:experiment-environments}. 
    }
\end{figure}

\changed{
Evident in the MPP results, \ddms can handle the frequent 
re-planning request of a few hundreds of robots in large environments. 
We further compare the \ddms variants in DMP and see how the 
benefits of heuristics are carried to the dynamic setting. 
Here, we do not involve other methods since the DMP formulation 
is relatively new and we could not locate comparable algorithms designed
for DMP under a warehouse setting in the research literature.
Our evaluation of DMP measures system throughput by the 
makespan for the robots to reach $10000$ uniformly randomly sampled 
goal configurations in total. 
Fig.~\ref{fig:results-dynamic} shows the evaluation results. 
The performance comparison between the \ddms variants is consistent 
with the one-shot results. 
The experiment also shows an interesting trend: the total 
makespan initially drops quickly as the number of robots increases; as 
the number of robots keeps increasing, the total makespan then begins to 
get larger again. This makes intuitive sense because too many robots are 
expected to make routing harder in more complex environments. 
The optimal makespan is achieved at around $150$ and $300$ robots. 
Viewing this together with Fig.~\ref{fig:results-static-warehouse}, we draw 
the conclusion that \ddms achieves a much faster computation speed with 
minor loss on optimality.
}


\changed{
All tests are repeated with varied grid sizes and 
obstacle percentages. The results, which are omitted due to space 
constraint, are consistent in all cases. 
}

A video of simulated \ddms runs can be found at \texttt{https://youtu.be/briO507tJiY}.

\section{Conclusion and Future Work}\label{sec:conclution}
In this work, we developed a decoupled multi-robot path planning 
algorithm, \ddms. With the proposed heuristics based on path 
diversification, which seeks to balance the use of graph vertices, 
and the employment of sub-problem solution databases for fast and 
optimal local conflict resolution, \ddms is empirically shown to 
achieve significantly faster computational speed while producing 
high quality solutions, for both one-shot and dynamic problem settings. 

In proposing \ddms, our hope is to optimize the algorithm for the 
two main phases of a decoupled approach. While the initial iteration
of \ddms shows promising performance, in future work, we would 
like to apply the novel heuristics from \ddms for solving  
multi-robot path planning problems beyond warehouse-like settings. 
In addition, improvements to these heuristics are possible.
%
For example, we only attempted $2\times 3$ and $3\times 3$ 
solution databases; databases using other graphs might provide 
better performance. 
Also, tighter integration of the two heuristics may further boost 
performance.


{\small
\bibliographystyle{IEEEtran}
\bibliography{../bib/all}
}

\end{document}